\pdfoutput=1

\documentclass[letterpaper]{article} 
\usepackage[pass]{geometry}
\usepackage{ijcai18}
\usepackage{booktabs}
\usepackage{todonotes}

\usepackage{times}
\usepackage{xcolor}
\usepackage{soul}
\usepackage[utf8]{inputenc}
\usepackage[small]{caption}
\usepackage{todonotes}

\title{Reinforcement Learning using Augmented Neural Networks}

\author{
Jack Shannon,
Marek Grzes
\\
School of Computing, University of Kent, UK \\
\{js975, m.grzes\}@kent.ac.uk
}

\begin{document}

\maketitle

\begin{abstract}
Neural networks allow Q-learning reinforcement learning agents such as deep Q-networks (DQN) to approximate complex mappings from state spaces to value functions. However, this also brings drawbacks when compared to other function approximators such as tile coding or their generalisations, radial basis functions (RBF) because they introduce instability due to the side effect of globalised updates present in neural networks. This instability does not even vanish in neural networks that do not have any hidden layers. In this paper, we show that simple modifications to the structure of the neural network can improve stability of DQN learning when a multi-layer perceptron is used for function approximation.
\end{abstract}

\section{Introduction}

Deep Reinforcement Learning, a group of algorithms that combine Reinforcement Learning and Deep Neural Networks, has proven to be successful in displays of human level artificial intelligence. Examples include Deep Q-Networks \cite{Mnih2015} which learnt to play Atari 2600 games, in some cases at a superhuman level, AlphaGo \cite{Silver2016} which beat the world's best player in the game Go, and IBM Watson DeepQA \cite{Ferrucci2010} which won against top human players in the TV gameshow Jeopardy. Due to the success in these domains and the nature of it being a trial and error approach, it seems that Deep Reinforcement Learning could be a forerunner to achieving human-level, general artificial intelligence, a machine that could match or exceed human performance at any task \cite{Lake2017}.

In contrast to classification/supervised learning which uses static data, Reinforcement Learning (RL) uses dynamic data which makes using Neural Networks challenging. By dynamic data, we mean that the required output of the model is constantly changing and evolving. Neural networks excel at classification when the data is static since it takes many iterative steps to train the parameters of the network. In reinforcement learning, however, the data mappings are not static. RL learns de novo (from the beginning), there is no stationary training data to learn from, experience is iteratively built up, and the learning algorithm continually samples from and updates the model. In the early research on the use of neural networks for function approximation in reinforcement learning, Reidmiller \cite{Riedmiller2005} had to invent techniques to account for the requirements of RL and stabilise learning. Specifically, Reidmiller proposed the solution of storing all previous observations and replaying them while training. This accounted for the instability where previously learnt information would be lost due to only training against the latest observation. However, this method is not feasible on large problems since updating the value function using replays from all previously explored states quickly becomes computationally expensive. The replay data is also highly correlated which causes problems when training a neural network.

Mnih \cite{Mnih2015} proposed the use of cloning the network and freezing the weights to read the values of the new states from a stable network. Experience Replay \cite{lin92selfimproving} was also randomly sampled in batches and used to train the network, which decorrelates the data. Distributed approaches to deep reinforcement learning have been used to successfully stabilise learning, by aggregating shared experiments with multiple agents running in parallel \cite{Mnih2016}. While this approach works well, it is not possible to use in all applications, such as robotics where it is impractical to have parallel agents. Deep Q-Networks (DQN) is one of the key Deep Reinforcement Learning frameworks that is state-of-the-art today, and the number of its applications is growing \cite{ghesu2016artificial}.

Using Neural Networks and training them with gradient descent introduces instability, since small changes in training data can produce very different models. An ensemble of neural networks that combines their results is one way to effectively mitigate this issue \cite{Cunningham2000}. Another key problem throughout Deep Learning research has been the vanishing gradient problem caused by sigmoidal activation functions. For this reason, Rectified Linear Unit (ReLU) activations are now widely used to counter this instability \cite{Nair2010} \cite{Lecun2015}. In Reinforcement Learning, Neural Networks with global basis functions \cite{bishop96nnbook} have the advantage of being scalable to large state spaces, and they are capable of estimating the value of unseen states. Unfortunately, due to this generalisation power, they also lose out on being able to make local updates. This is because one batch of data samples presented to the neural networks learning algorithm can change the values of distant states that are far from any state in the batch. For the same reason, the global updates can also have impact on classical RL concepts such as optimistic initialisation which can guide exploration.

Other reinforcement learning function approximators are often a good choice, such as tile coding \cite{Sutton1998} or Radial Basis Function Networks (RBFN) \cite{Orr1996}. These have the advantage of being less prone to overwriting previously learnt data because tile coding and RBFNs are able to update local areas of the value function while preserving global values, allowing them to take advantage of optimistic initialisation, an exploration strategy that prioritises visiting unseen states \cite{Sutton1998,brafman02rmax}. However, in spite of these advantages, both tile coding and RBFNs suffer from the curse of dimensionality and they cannot be scaled up to huge problems that can still be modelled using Multi-Layer Perceptrons (MLPs) \cite{bertsekas96neuro-dynamic}. Therefore, our research on MLPs in the context of reinforcement learning is well-justified because MLPs scale up well, and we show how to make them more effective in reinforcement learning applications.


The main contribution of this paper aims at providing a new way of stabilising learning when using neural networks based on a multi-layer perceptron. We aim to take advantage of the positive properties of using neural networks in RL and eliminate certain negative properties which inhibit stable learning.

In this paper, we introduce the use of Square Multi-Layer Perceptrons (SMLPs) \cite{Flake2012} in Deep Reinforcement Learning, as a tool to make the process stable, and therefore speed up learning. The most successful way to use neural networks in RL is a DQN approach, so we will be taking that algorithm and applying this technique. In Section 2, we discuss the necessary background knowledge. In Section 3, we present the algorithms used. In Section 4, we show the methodology used in our experiments and the domains which will be tested. Section 5 will be a discussion of the results.



\section{Background}


The agent interacts with the environment that transmits a state and a reward. The goal of Reinforcement Learning is for the agent to learn an optimal policy for choosing an action which maximises the total reward that can be received in expectation.


The Reinforcement Learning problem can be formalized as the Optimal Control of a Markov Decision Process (MDP). An MDP is a tuple $< S,A,P,R,\gamma >$ where $S$ is a set of all states, $A$ is a set of all actions, $P$ is a transition function that returns the probability of the next state being $s'$ if we take action $a$. $R$ is a reward function that returns a reward value $r$ given as a result of taking action $a$ in state $s$. Finally, $\gamma$ is the discount factor and defines the difference in importance of present rewards vs future rewards. A policy $\pi$ defines the behaviour of the agent.


We can learn optimal policies for MDPs using temporal difference methods such as Q-learning, which use dynamic programming techniques, iteratively making predictions and updating the policy using observations. This iterative process is required in learning problems where reward feedback on an action is not always immediate, and it may be seen many timesteps later.
%

Q-learning learns an \textit{action-value} function. Given a state and an action, it returns the value of taking that action. The Q-values are updated using the following rule, usually performing the update in an online manner, sampling directly from the environment each timestep:
\begin{equation}
Q(s,a) \leftarrow Q(s,a)+\alpha [ r+\gamma  \max_{a'}{Q(s',a')} - Q(s,a)].\label{eq:qupdate}
\end{equation}

The Q-learning update rule (Equation \ref{eq:qupdate}), updates the current estimate of the Q-value by calculating the sum of the immediate reward and the discounted future reward. This discounted future reward is estimated by taking the maximum Q-value of all possible actions in the next state, and multiplying by the discount factor $\gamma$. The Q-learning update rule gives a new estimate which is used to update the stored Q-value by taking the difference and updating proportionally to the learning rate $\alpha$.


With Q-learning the policy can be derived greedily by choosing the action with the maximum Q-value at a state: $\max_{a}{Q(s,q)}$. Typically, during learning we use an $\epsilon$-greedy policy, which uses a random exploration probability $\epsilon$, and otherwise a greedy policy will be used. We choose the action with the largest Q-value, and otherwise it will choose a random action. In practice, linearly annealing $\epsilon$ is often used which allows the agent to use lots of random exploration at the start of learning and less later on.


The Q-function can be represented using a look up table. As the state space becomes larger, lookup tables require an unreasonable amount of memory and become infeasible. It can also be appropriate to approximate states with similar features to deal with the curse of dimensionality and to speed up learning. Fortunately, a function approximator can allow all states to be represented, event if it is impossible to enumerate all states. Tile coding \cite{Sutton1998} is one form of linear function approximator, which discretises continuous representations of state, and Radial Basis Functions are a generalisation of this. We can also use multi-layer perceptrons to represent value functions.

A common reinforcement learning exploration strategy is known as optimistic initialisation \cite{Sutton1998}. Here we let the agent assume that unseen states are more valuable, by initialising the value function in such a way that initial values are higher than their true value. This means that the agent will prefer to take an action of an unseen state where the expected value is high. Unfortunately, function approximation with global basis functions, e.g., multi-layer perceptrons, will normally destroy this initialisation after a few updates of the parameters in the network, and the RL algorithms cannot benefit from optimistic exploration in such situations. However, the localised updates that we investigate in this paper may better preserve optimism.




\section{Algorithms}

This section introduces main technical details of the algorithms investigated in this paper.

\subsection{Deep Q-Networks}
Deep Q-Networks (DQN) is a Q-learning algorithm that uses Neural Networks as a function approximator. The main elements of the DQN algorithm are presented below.

Given a state, actions are selected using the Q-function which is approximated by a neural network. This is repeated for every timestep until the episode is complete. An episode is complete after a maximum number of iterations or until a goal state is reached. Each iteration, the agent receives an observation from the environment. An observation consists of the state, action taken, next state, reward and  whether it is a terminal state. This is saved into the Experience Replay Buffer. The buffer records these observations as the episode is played through. For training the network, observations are randomly sampled in batches. The parameters of the neural network are updated by randomly sampling a batch from the experience replay buffer and training the network using back-propagation. The Q-learning update rule (Equation \ref{eq:qupdate}) is used on our batch to generate our training data, reading $Q(s',a')$ from the target network. The mean squared error of the difference between the target value and the current value becomes a loss function, and the RMSProp Optimizer is generally used for batch learning.

With supervised learning, targets are fixed before learning. With reinforcement learning, this is not the case and targets are derived by reading from the current/target model. This leads to instability while learning when using a neural network.

DQN with frozen target networks provides a more stable model to read $Q(s',a')$ from. At specified intervals, the target network clones and fixes weights from the main model \cite{Mnih2015}.

\subsection{Square Multi-Layer Perceptron}

A Multi-Layer Perceptron (MLP) is a feed forward neural network with one or more hidden layers. The output $y$ of a simple MLP with one hidden layer with a sigmoidal activation function $\phi$, input $x$ and with weight vectors $v$ and $w$ is shown:
\begin{equation}
y = \sum_j {v_j\: \phi \left ( \sum_i w_{ji}x_i + b_j \right )}.\label{eq:mlp}
\end{equation}

A Radial Basis Function Network (RBFN) is a neural network that uses a radial basis function (RBF) as activation functions \cite{bishop96nnbook}. A radial basis function is a real-valued function whose value depends on the distance from the origin, typically Gaussian with the parameters $\mu_{ji}$ and $\sigma$. The output $y$ of an RBF network with one hidden layer is shown:
\begin{equation}
y = \sum_j {v_j\: e^{-\frac{1}{2} \sum_i \left ( \frac{x_i - \mu _{ji}}{\sigma } \right )^2}}.\label{eq:rbfn}
\end{equation}

The main advantage of using an RBF as the activation function in a neural network is its ability for the function to be more spatially localised. However, RBFNs have the disadvantage of suffering from the curse of dimensionality, whereas MLPs do not suffer as much.



Fortunately, we can augment an MLP to behave like an RBFN, and still cope well with the curse of dimensionality. A Square Multi-Layer Perceptron (SMLP) \cite{Flake2012} takes an MLP and provides it with a second set of inputs, which are squared copies of the original inputs. One way to implement this is to take an MLP with an input pattern of $n$ components and extend the input pattern so that it is now $2n$. Then as part of a pre-processing step, we duplicate our input data, and each value of the duplicated input data is squared. This duplicated input data is then joined with the original input data and passed to the neural network as one input vector.

This can also be implemented by changing the architecture of an MLP (Equation \ref{eq:mlp}) to support a second set of weights $u$ for the squared inputs. The output $y$ of an SMLP is shown:
\begin{equation}
y = \sum_j {v_j\: \phi \left ( \sum_i w_{ji}x_i + u_{ji}x_i^2 + b_j \right )}.\label{eq:smlp}
\end{equation}

Flake showed that supplying an MLP with duplicated and squared inputs gives an MLP the capability of approximating an RBFN, and allows it to share both intrinsic properties. The combination of a sigmoidal activation function $\phi$ and the squaring of the duplicated inputs was analytically shown to be approximately equivalent to an RBFN network. Flake showed that this augmentation of an MLP allows both a global and local representation of the input space. Local artefacts can be preserved whilst not influencing global areas of the input space. Flake presented good results using SMLPs on a complex classification problem. The function learnt shows familiar traits that we have seen in reinforcement learning value functions.

\subsection{Our contribution}

We propose that a SMLP will outperform an MLP, when used as a function approximator for Reinforcement Learning algorithms. This is because local updating will lead to more stable exploration of the environment. Additionally, the RL agent can maintain optimistic initialisation due to the properties of an RBF being able to represent spatially localised features. An SMLP has been shown to represent complex functions more easily due to more complex decision boundaries learnt. SMLPs should also still perform well on problems which require extreme generalisation across unseen states, since SMLPs have a dual ability to behave as both an RBFN or an MLP. All in all, SMLPs are particularly suitable for function approximation in reinforcement learning.

In our experiments, we will use an SMLP with the DQN algorithm, to see if we are able to use the benefits of both an RBF Network and an MLP. We propose that SMLPs may work particularly well when used as an approximation function in RL because due to their local nature they are a more suitable architecture for representing Q-functions.

\section{Methodology}

This section outlines the domains used to evaluate our contribution. We test our algorithm on two classic control problems from the Reinforcement Learning literature, Mountain Car and Acrobot. We also designed a spiral-shaped maze for a Gridworld problem seen in the Reinforcement Learning literature \cite{asmuth08potential}, which would require the use of a Q-function with a similar shape to the function learnt in a double spiral classification problem. For each domain we describe the problem, determine the solution, and describe the state representation and the reward function.

\subsection{Mountain Car}

The Mountain Car problem \cite{Sutton1998} features a car with limited engine power, meaning that it is unable to drive directly to the top of a steep valley using full throttle. The solution is for the car to oscillate left and right to build the inertia necessary to reach the goal at the top right. The state representation is the x position and the velocity which is clipped. The reward function given is -1 at all steps until it reaches the goal at the top right where it receives a reward of 0 and the episode is terminated.

\subsection{Acrobot}

Acrobot is a freely moving double arm pendulum. The task is to apply torque to the joint between the arms and swing the end of the pendulum above a height of the length of one of the arms. The state representation is the joint angles and their velocities. The reward function given is -1 at all timesteps until it reaches the goal where it receives a reward of 0 and the episode is terminated.

\subsection{Spiral Maze Gridworld}

\begin{figure}
\centering
\includegraphics[width=.3\textwidth]{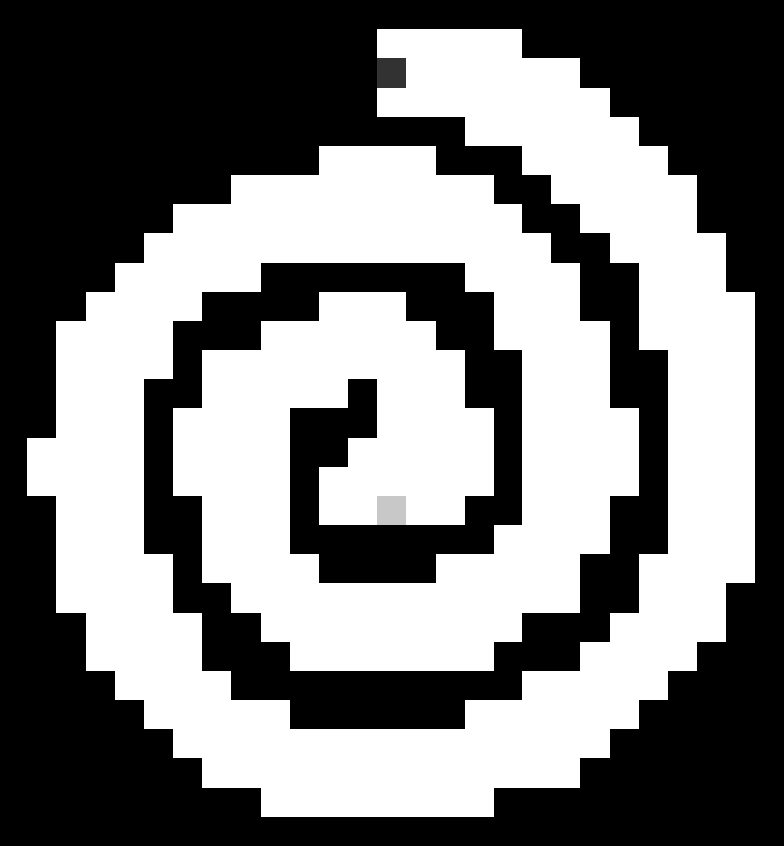}
\caption{The spiral maze used in the gridworld problem. The goal state is in the centre of the maze, the start state is at the entrance to the maze.}
\label{SpiralMazeImg}
\end{figure}

Taking inspiration from \cite{Flake2012}, who used the double spiral problem as an experiment to show the effectiveness of SMLP, we chose a spiral shaped maze (Figure \ref{SpiralMazeImg}) and turned it into a Gridworld problem. The agent will have to navigate the maze and reach the centre with just two features, the x/y position coordinates as state representation.

In this environment, the agent is in a 2D grid with walls blocking its path. The task is to choose from one of four directions to travel in at each timestep. The environment is made stochastic by adding a slip probability of 0.2 which will cause the agent to travel in a random unintended direction. The reward function given is -1 at all steps until it reaches the goal at centre of the maze where it receives a reward of 0 and the episode is terminated.

A table based Q-learning agent with a sufficiently fine discretisation of the variables x/y can learn this task easily. Thanks to optimistic initialisation, the agent will value unseen states higher and rapidly explore. However, this task is particularly difficult to learn with an MLP which doesn't always preserve optimistic initialisation, so often relies on a larger amount of samples from the environment and random exploration. By testing on this domain, we may see evidence of better exploration and an indication that SMLPs will take advantage of optimistic initialisation.

%

\subsection{Experiment Setup}

We use DQN along with experience replay and a target network to stabilise learning, as seen in \cite{Mnih2015}. The experience replay uses a memory size of 300,000 samples and the target network is updated every 1000 timesteps. We began each experiment with random exploration to fill experience replay; no training was done during this period. After this, an $\epsilon$-greedy exploration policy was used with $\epsilon$ being linearly annealed over 1000 timesteps.
The discount factor $\gamma$ was equal to 0.95. We used an MLP architecture with 2 hidden layers of 256 neurons with ReLU activations, which was empirically chosen. The output layer using a linear activation. We used the RMSProp optimiser for training the networks, which is a specialised gradient descent training algorithm for batched learning.

Prior to training, the weights of the neural network are initialised close to zero which means that the initial value output for all states will be close to zero. Also, our reward functions in all domains are set up so that a negative reward is awarded for all moves except for the final move to the goal state. Since our initial value function returns a higher value than the true value function, our agent is optimistically initialised. We should add that optimistic exploration is beneficial in all our testing domains because specific, non-random behaviour is required to reach their goal states. We know from the literature that random exploration may require an exponential number of steps to reach the goal state for the first time \cite{thrun92efficient}.

The total reward that the agent received is summed over an episode. We are comparing our algorithm against a standard MLP without squared inputs but with identical settings. We are comparing the total reward per episode in which higher is better. This is shown in the graphs with a red line for when SMLP is used as the function approximator, and a blue line when using an MLP. Each experiment is averaged over 10 trials and the results are shown using a moving average to smooth out the line on the graph. We also display error bars of the standard error of the mean (SEM) to visualise  variation in the results between different instances of the experiments.

\section{Results}

We are considering two different indicators to compare the relative performance of SMLP and MLP. One of these is the long-term behaviour which indicates the degree to which the algorithm is learning the problem at all. And secondly, we take into account how fast it does so, that is, how many training episodes are required to reach a certain performance.

\begin{figure}
\centering
\includegraphics[width=.4\textwidth]{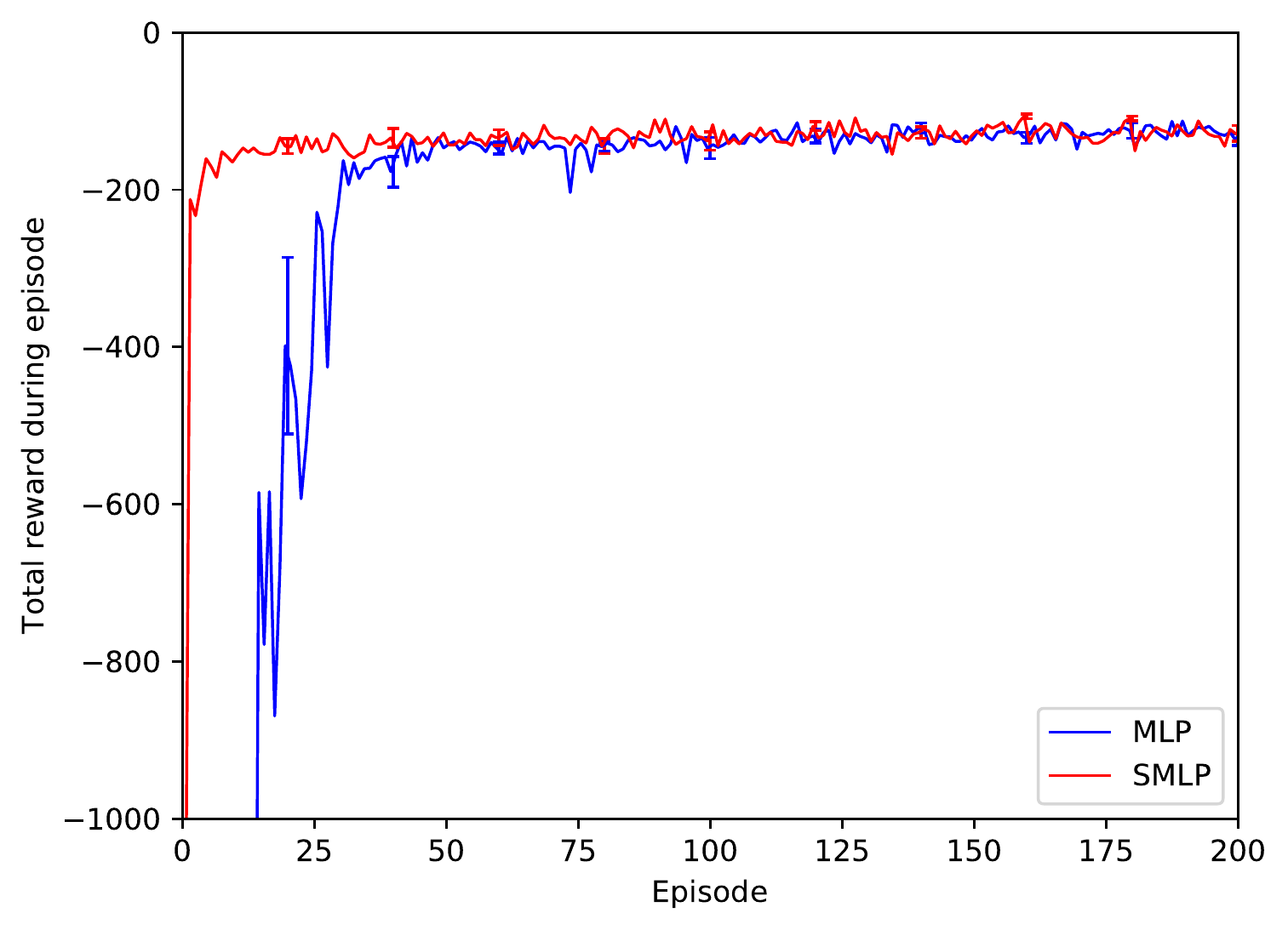}
\caption{Mountain Car. Comparing total reward received per episode with and without squared inputs. SMLP in red, MLP in blue.}
\label{MountainCar}
\end{figure}

\begin{figure}[t]
\centering
\includegraphics[width=.4\textwidth]{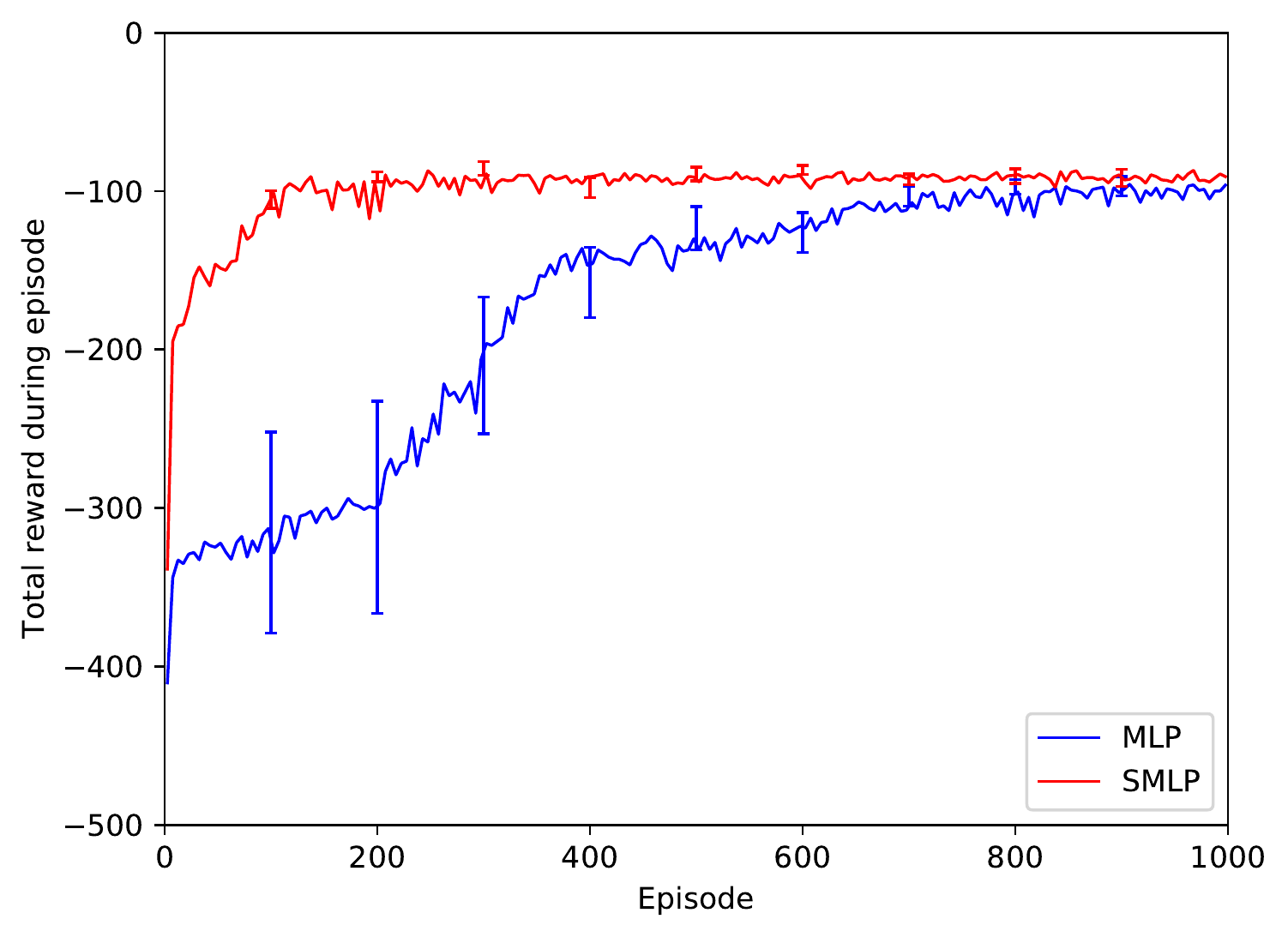}
\caption{Acrobot. Comparing total reward received per episode with and without squared inputs. SMLP in red, MLP in blue.}
\label{Acrobot}
\end{figure}

\begin{figure}
\centering
\includegraphics[width=.4\textwidth]{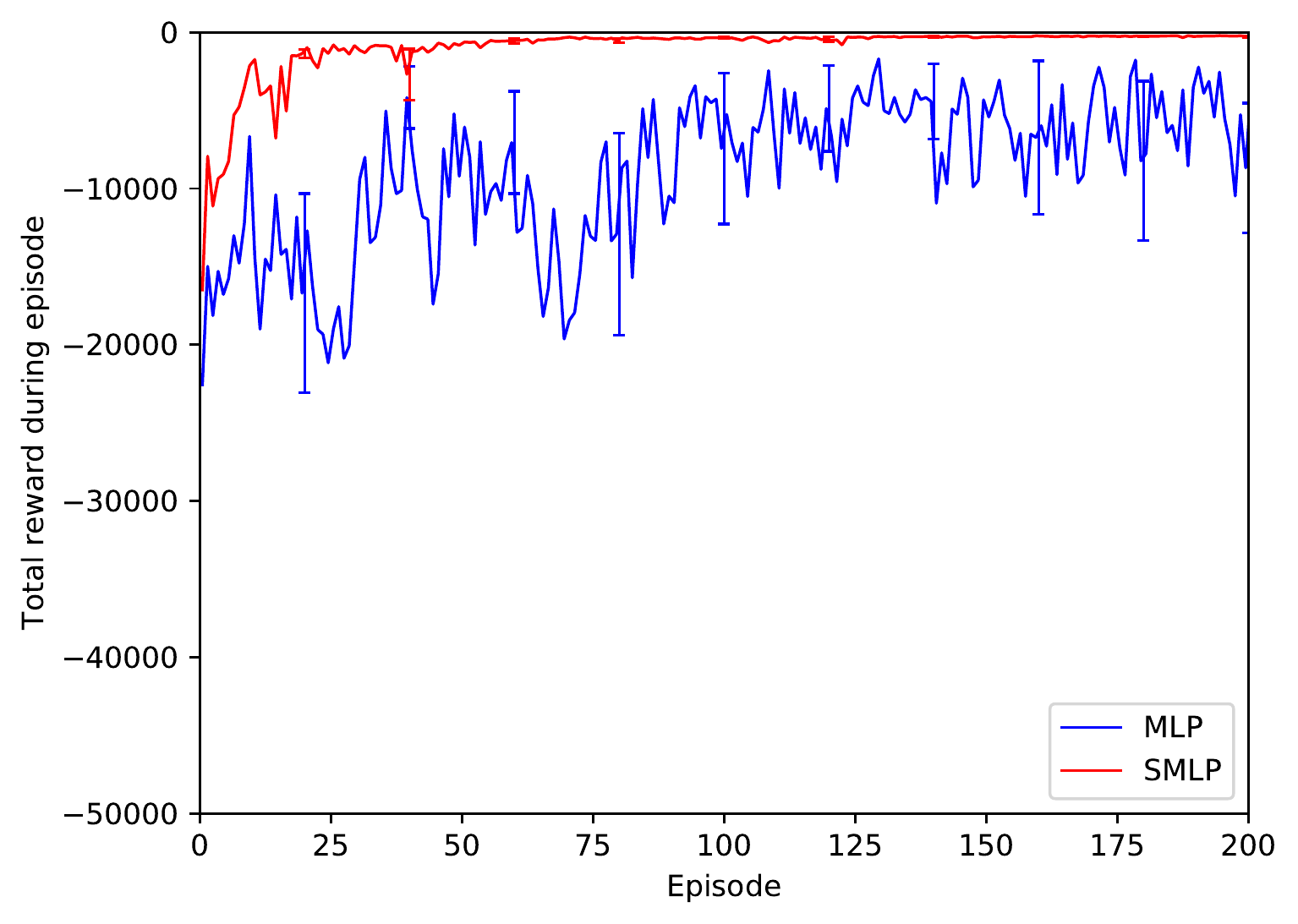}
\caption{Spiral Maze Grid World. Comparing total reward received per episode with and without squared inputs. SMLP in red, MLP in blue.}
\label{Spiral}
\end{figure}

We show that in these 3 examples, the SMLP is consistently outperforming the standard MLP. By this we mean that it learns faster, i.e. in fewer episodes, and in some cases, it also achieves a better long-term reward. There are differences however, between the individual case studies.

In the case of the spiral maze gridworld (Figure \ref{Spiral}), the MLP does not learn the problem as well as the SMLP within the 200 episodes. The reason for that is the complex structure of the value function that is similar to the function learned by Flake in his paper \cite{Flake2012}.

On the other hand, in the Acrobot (Figure \ref{Acrobot}) and the Mountain Car (Figure \ref{MountainCar}) domains, the early performance of the MLP is also poor compared to the SMLP, but the in the long run it learns the problem as well as the SMLP. Indeed, for the Mountain Car, the total reward per episode is indistinguishable for both networks after 50 episodes. These results indicate that the structure of the value function in the Acrobot and Mountain Car domains is simple enough for the MLP to represent so that it can be learned by the MLP when a sufficient amount of exploration is guaranteed. However, considering efficiency of exploration, SMLP excels in these experiments because localised updates lead to faster convergence and exploration that is based on the value function improves faster.

Common to all examples is that the MLP solutions take much longer to reach the same learning objective as the SMLP. Particularly in the Acrobot, the increase of the total reward is slow on the MLP. And, similarly, the Mountain Car it takes many more episodes to reach the same total reward.

Overall, our empirical evidence shows that SMLP can improve the asymptotic convergence when the value function is complex (spiral maze gridworld), and it can improve exploration regardless of the complexity of the value function as long as exploration is challenging in a given domain (mountain car and acrobot).

There is a clear instability during MLP learning. This could be due to the failure of an MLP to represent the local features of a spiral maze, whereas SMLPs have the ability to form more complex decision boundaries. A similar result was seen on the double spiral maze example in \cite{Flake2012}, where an MLP found the domain challenging. This follows the established result that SMLPs can approximate certain functions with less training, even though the end result may be the same.

There are at least two possible reasons for the difference in performance between SMLP and MLP. Firstly, it's an established result by the paper by Flake, where it was shown that squared inputs require less training iterations to learn than MLP. Note that SMLP has more parameters than MLP, yet it can learn with fewer data examples than MLP. A second explanation could be related to the local vs global update properties, with SMLP presumably being more local in its updating than MLP. The impact of this feature may be highly important in reinforcement learning. In particular, our knowledge of exploration in reinforcement learning indicates that more consistent value functions due to localised updates can lead to more informed learning of the environment and an associated policy to control it.

A perfect local update is when only the intended point in the state space is updated by changing the weights, and no other point in the state space is affected. This happens when the value function is represented in a tabular form. However, to speed up learning and benefit from generalisation, reinforcement learning practitioners normally want the updates to be shared with neighbouring states that have not been visited. Tile coding or RBFNs satisfy this requirement and the updates of a particular state do not affect the Q-function for states that are `very different' from the updated state. This property makes tile coding and RBFNs very powerful on problems with relatively small state spaces. In general, neural networks with global basis functions, e.g., MLPs, are useful in reinforcement learning as they cope with the curse of dimensionality, which is a challenge for local basis functions. Our results show that a middle ground can be found, and adding a localised architecture to MLPs is an advantage in reinforcement learning due to its dynamic nature. In domains where we have a large state space, an MLP may be preferred or even required, and our results show that its learning can be improved, i.e., a good result can be achieved after a smaller amount of exploration or even a better asymptotic solution can be found.

In all our experiments, SMLPs outperform and show faster convergence than MLPs. This quick convergence shows that the agent has a more developed Q-function earlier on, perhaps due to the localisation of a SMLP and resulting stability. SMLPs are also capable of a more complex set of decision boundaries than both MLPs and RBF networks, as outlined in \cite{Flake2012}. The results imply that MLPs have a fair amount of difficulty in learning a correct Q-function approximation for these problems. Whereas SMLPs quickly learn a Q-function that gives an optimal policy. The error bars show that using SMLP as a function approximator is stable across trials.


%



\section{Conclusion}
This paper introduced using a Square-MLP neural network as a function approximator for the deep reinforcement learning algorithm DQN. Our method achieves faster convergence than an MLP in all tested domains. We have demonstrated that our algorithm outperforms MLPs in domains that benefit from optimistic initialisation. Strategies for exploration have long been an important design choice when developing reinforcement learning agents. Using neural networks with global basis functions as an approximation function destroys optimistic initialisation that forms an important exploration strategy in reinforcement learning. We have shown that squaring the inputs to the neural network can hope to mitigate this.

In future work, we will look at the performance of these networks on larger problems, to see the scalability of this approach. Since such a simple modification on the architecture of a neural network has the capability to dramatically improve the performance of a DQN agent, this method could be a useful tool for Deep Reinforcement Learning researchers.

\bibliographystyle{named}
\bibliography{bibliography}  

\begin{thebibliography}{}

\bibitem[\protect\citeauthoryear{Asmuth \bgroup \em et al.\egroup
  }{2008}]{asmuth08potential}
John Asmuth, Michael~L. Littman, and Robert Zinkov.
\newblock Potential-based shaping in model-based reinforcement learning.
\newblock In {\em Proceedings of {AAAI}}, 2008.

\bibitem[\protect\citeauthoryear{Bertsekas and
  Tsitsiklis}{1996}]{bertsekas96neuro-dynamic}
Dimitri~P. Bertsekas and John~N. Tsitsiklis.
\newblock {\em Neuro-Dynamic Programming}.
\newblock Athena Scientific, 1996.

\bibitem[\protect\citeauthoryear{Bishop}{1996}]{bishop96nnbook}
Christopher~M. Bishop.
\newblock {\em Neural Networks for Pattern Recognition}.
\newblock Oxford University Press, 1996.

\bibitem[\protect\citeauthoryear{Brafman and Tennenholtz}{2002}]{brafman02rmax}
Ronen~I. Brafman and Moshe Tennenholtz.
\newblock R-max - a general polynomial time algorithm for near-optimal
  reinforcement learning.
\newblock {\em {JMLR}}, 3:213--231, 2002.

\bibitem[\protect\citeauthoryear{Cunningham \bgroup \em et al.\egroup
  }{2000}]{Cunningham2000}
P{\'a}Draig Cunningham, John Carney, and Saji Jacob.
\newblock Stability problems with artificial neural networks and the ensemble
  solution.
\newblock {\em Artificial Intelligence in medicine}, 20(3):217--225, 2000.

\bibitem[\protect\citeauthoryear{Ferrucci \bgroup \em et al.\egroup
  }{2010}]{Ferrucci2010}
David Ferrucci, Eric Brown, Jennifer Chu-Carroll, James Fan, David Gondek,
  Aditya~A Kalyanpur, Adam Lally, J~William Murdock, Eric Nyberg, John Prager,
  et~al.
\newblock Building watson: An overview of the deepqa project.
\newblock {\em AI magazine}, 31(3):59--79, 2010.

\bibitem[\protect\citeauthoryear{Flake}{2012}]{Flake2012}
Gary~William Flake.
\newblock Square unit augmented, radially extended, multilayer perceptrons.
\newblock In {\em Neural networks: tricks of the trade}, pages 143--161.
  Springer, 2012.

\bibitem[\protect\citeauthoryear{Ghesu \bgroup \em et al.\egroup
  }{2016}]{ghesu2016artificial}
Florin~C Ghesu, Bogdan Georgescu, Tommaso Mansi, Dominik Neumann, Joachim
  Hornegger, and Dorin Comaniciu.
\newblock An artificial agent for anatomical landmark detection in medical
  images.
\newblock In {\em International Conference on Medical Image Computing and
  Computer-Assisted Intervention}, pages 229--237. Springer, 2016.

\bibitem[\protect\citeauthoryear{Lake \bgroup \em et al.\egroup
  }{2017}]{Lake2017}
Brenden~M Lake, Tomer~D Ullman, Joshua~B Tenenbaum, and Samuel~J Gershman.
\newblock Building machines that learn and think like people.
\newblock {\em Behavioral and Brain Sciences}, 40, 2017.

\bibitem[\protect\citeauthoryear{LeCun \bgroup \em et al.\egroup
  }{2015}]{Lecun2015}
Yann LeCun, Yoshua Bengio, and Geoffrey Hinton.
\newblock Deep learning.
\newblock {\em Nature}, 521(7553):436, 2015.

\bibitem[\protect\citeauthoryear{Lin}{1992}]{lin92selfimproving}
Long-Ji Lin.
\newblock Self-improving reactive agents based on reinforcement learning,
  planning and teaching.
\newblock {\em {M}achine {L}earning}, 8:293--321, 1992.

\bibitem[\protect\citeauthoryear{Mnih \bgroup \em et al.\egroup
  }{2015}]{Mnih2015}
Volodymyr Mnih, Koray Kavukcuoglu, David Silver, Andrei~A. Rusu, Joel Veness,
  Marc~G. Bellemare, Alex Graves, Martin Riedmiller, Andreas~K. Fidjeland,
  Georg Ostrovski, et~al.
\newblock Human-level control through deep reinforcement learning.
\newblock {\em Nature}, 518(7540):529--533, 2015.

\bibitem[\protect\citeauthoryear{Mnih \bgroup \em et al.\egroup
  }{2016}]{Mnih2016}
Volodymyr Mnih, Adria~Puigdomenech Badia, Mehdi Mirza, Alex Graves, Timothy
  Lillicrap, Tim Harley, David Silver, and Koray Kavukcuoglu.
\newblock Asynchronous methods for deep reinforcement learning.
\newblock In {\em International Conference on Machine Learning}, pages
  1928--1937, 2016.

\bibitem[\protect\citeauthoryear{Nair and Hinton}{2010}]{Nair2010}
Vinod Nair and Geoffrey~E Hinton.
\newblock Rectified linear units improve restricted boltzmann machines.
\newblock In {\em Proceedings of the 27th international conference on machine
  learning (ICML-10)}, pages 807--814, 2010.

\bibitem[\protect\citeauthoryear{Orr}{1996}]{Orr1996}
Mark J.~L. Orr.
\newblock Introduction to radial basis function networks, 1996.

\bibitem[\protect\citeauthoryear{Riedmiller}{2005}]{Riedmiller2005}
Martin Riedmiller.
\newblock Neural fitted q iteration -- first experiences with a data efficient
  neural reinforcement learning method.
\newblock In {\em Proceedings of the 16th European Conference on Machine
  Learning}, ECML'05, pages 317--328, Berlin, Heidelberg, 2005.
  Springer-Verlag.

\bibitem[\protect\citeauthoryear{Silver \bgroup \em et al.\egroup
  }{2016}]{Silver2016}
David Silver, Aja Huang, Chris~J Maddison, Arthur Guez, Laurent Sifre, George
  Van Den~Driessche, Julian Schrittwieser, Ioannis Antonoglou, Veda
  Panneershelvam, Marc Lanctot, et~al.
\newblock Mastering the game of go with deep neural networks and tree search.
\newblock {\em Nature}, 529(7587):484--489, 2016.

\bibitem[\protect\citeauthoryear{Sutton and Barto}{1998}]{Sutton1998}
Richard~S. Sutton and Andrew~G. Barto.
\newblock {\em Introduction to Reinforcement Learning}.
\newblock MIT Press, Cambridge, MA, USA, 1st edition, 1998.

\bibitem[\protect\citeauthoryear{Thrun}{1992}]{thrun92efficient}
S.~Thrun.
\newblock Efficient exploration in reinforcement learning.
\newblock Technical Report CMU-CS-92-102, Carnegie Mellon University, Computer
  Science Department, 1992.

\end{thebibliography}


\end{document}